\RequirePackage{fix-cm}
\documentclass[natbib,twocolumn]{svjour3}          
\smartqed  
\usepackage{graphicx}
\usepackage{microtype}
\usepackage{xspace}
\usepackage{algorithm}
\usepackage{algorithmicx}
\usepackage{algpseudocode}

\usepackage{booktabs}
\usepackage{tabularx}
\newcolumntype{Y}{>{\centering\arraybackslash}X}
\usepackage{xspace}
\usepackage{esvect}
\usepackage{xcolor}
\usepackage[pagebackref=true,breaklinks=true,citecolor=tealblue,colorlinks,linkcolor=ruby,bookmarks=false]{hyperref}
\definecolor{ruby}{rgb}{0.88, 0.07, 0.37}
\definecolor{tealblue}{rgb}{0.18, 0.40, 0.46}
\usepackage{array}
\usepackage{multirow}
\newcolumntype{H}{>{\setbox0=\hbox\bgroup}c<{\egroup}@{}}
\usepackage{amsmath}
\usepackage{amsfonts}
\begin{document}

\title{RIConv++: Effective Rotation Invariant Convolutions for 3D Point Clouds Deep Learning}


\author{Zhiyuan Zhang         \and
        Binh-Son Hua          \and
        Sai-Kit Yeung
}


\institute{Zhiyuan Zhang \at
              Ningbo Research Institute, Zhejiang University \\
              College of Information Science and Electronic Engineering, Zhejiang University\\
              NingboTech University\\
              \email{cszyzhang@gmail.com}           
           \and
           Binh-Son Hua \at
              VinAI, Vietnam\\
              \email{binhson.hua@gmail.com}
           \and
           Sai-Kit Yeung \at
              Hong Kong University of Science and Technology\\
              \email{saikit@ust.hk}
}

\date{Received: date / Accepted: date}

\def\ourconv{RIConv++\xspace}
\def\smallgap{\vspace{0.05in}}
\maketitle

\begin{abstract}
3D point clouds deep learning is a promising field of research that allows a neural network to learn features of point clouds directly, making it a robust tool for solving 3D scene understanding tasks. 
While recent works show that point cloud convolutions can be invariant to translation and point permutation, investigations of the rotation invariance property for point cloud convolution has been so far scarce. 
Some existing methods perform point cloud convolutions with rotation-invariant features, existing methods generally do not perform as well as translation-invariant only counterpart.
In this work, we argue that a key reason is that compared to point coordinates, rotation-invariant features consumed by point cloud convolution are not as distinctive.
To address this problem, we propose a simple yet effective convolution operator that enhances feature distinction by designing powerful rotation invariant features from the local regions. 
We consider the relationship between the point of interest and its neighbors as well as the internal relationship of the neighbors to largely improve the feature descriptiveness. 
Our network architecture can capture both local and global context by simply tuning the neighborhood size in each convolution layer. 
We conduct several experiments on synthetic and real-world point cloud classifications, part segmentation, and shape retrieval to evaluate our method, which achieves the state-of-the-art accuracy under challenging rotations.
\keywords{3D Point Cloud \and Convolutional Neural Networks \and Deep Learning \and Rotation Invariance}
\end{abstract}

\section{Introduction}
\label{intro}
3D scene understanding is a challenging problem in computer vision. 
With the wide availability of consumer-grade RGB-D and LiDAR sensors, acquiring 3D scenes has become easier, cheaper, resulting in many mid- and large-scale 3D datasets~\citep{wu-3dshapenets-cvpr15,chang2015shapenet,hua2016scenenn,dai2017scannet,armeni20163d,yi2016scalable,uy-scanobjectnn-iccv19}. 
One of the popular representations of such 3D data is the 3D point cloud representation. 
The recent advances of deep learning with 3D point clouds has led to opportunities to revisit and tackle 3D scene understanding from a new perspective.

The basic idea of 3D point clouds deep learning is to let a neural network consume a point cloud directly.
A point cloud is a mathematical set and so it fundamentally differs from an image, rendering traditional neural networks unsuitable for 3D point clouds.  
It is therefore necessary to design a convolution-equivalent operator in the 3D domain that can take a point cloud as input and output its per-point features. 
In the past few years, significant efforts have been made with promising results along this direction~\citep{qi2017pointnet,qi2017pointnet++,hua2017point,li2018pointcnn,xu2018spidercnn,zhang-shellnet-iccv19,zhao2020quaternion}.

Despite such research efforts, a property often overlooked in point cloud convolution is rotation invariance. 
This property arises from the fact that 3D data can have three degrees of freedom for rotation, making rotation invariance more challenging to achieve.
In the 2D domain, a viable solution is to augment the training data with random rotations. 
However, in 3D, such data augmentation becomes less effective due to the additional degrees of freedom, which can make training prohibitively expensive.

Some previous works have been proposed to learn rotation-invariant features~\citep{zhang2020global,zhang-riconv-3dv19,rao-spherical-cvpr19,poulenard-spherical-3dv19,deng2018ppf,chen2019clusternet}, which leads to consistent predictions given arbitrarily rotated point clouds. 
We observe that state-of-the-art methods can improve the feature learning by using local reference frame (LRF) to encode both local and global information~\citep{zhang2020global, kim2020rotation, thomas2020rotation}. 
However, LRF usually suffers sign flipping problem in the $x$ and $y$ axes, which makes rotation-invariant convolutions built upon LRF yield features not as distinctive as a translation-invariant convolution does. 
This can be demonstrated in the result of the object classification task: performing classification with aligned 3D shapes (using translation-invariant convolution) is more accurate than performing the same task with shapes with arbitrary rotations (using rotation-invariant convolution). 
For example, state-of-the-art methods with rotation invariance~\citep{zhang-riconv-3dv19,poulenard-spherical-3dv19,zhang2020global,kim2020rotation} reported classification accuracies about 86\%--89\% on ModelNet40~\cite{wu-3dshapenets-cvpr15} while methods without rotation invariance can reach accuracy as high as $93\%$~\citep{wang2018edgeconv,zhang-shellnet-iccv19}. This motivates us to address the limitation of the LRF and design a convolution that gives more informative features to increase the overall performance.

Particularly, we propose an effective and lightweight approach to perform rotation-invariant convolution for point clouds, which is an extension of our previous work~\citep{zhang-riconv-3dv19,zhang2020global}. 
We propose to make rotation-invariant features more \emph{informative} by local reference axis (LRA), and consider point-point relations, which improves feature distinction as well. Compared to LRF, we show that our LRA is more stable. To the best of our knowledge, our method performs consistent predictions across rotations, and reduces the performance gap compared to translation-invariant convolutions. 

In summary, the main contributions of this work are:
\begin{itemize}
\item \ourconv, an enhanced version of our previous convolution RIConv~\citep{zhang-riconv-3dv19}. 
We leverage local reference axis to achieve a stable rotation-invariant representation. We extract \emph{informative} rotation invariant features by considering the relationship between interest points and the neighbors as well as the internal relationship of the neighbors; 

\item A neural network architecture that stacks \ourconv for learning rotation-invariant features for 3D point clouds. The network can sense local, semi-global and global context by simply adjusting the neighborhood size and achieves consistent performance across different rotation scenarios;

\item Extensive experiments of our method on object classification, object part segmentation and shape retrieval that achieve the state-of-the-art performance under challenging scenarios including an analysis of rotation-invariant features and an ablation study of our neural network.
\end{itemize}

\section{Related Works}
\label{related}
3D deep learning began with a focus on regular and structured representations of 3D scenes such as multiple 2D images~\citep{su2015multi,qi2016volumetric,esteves2019equivariant}, 3D volumes \citep{qi2016volumetric,li2016fpnn}, hierarchical data structures like octree \citep{riegler2017octnet} or kd-trees \citep{klokov2017escape,wang2017cnn}. Such representations yield good performance, but they face challenges from a practical point of view: they require large memory consumption, have imprecise representation, and are not scalable to high-resolution data. Many recent works in 3D deep learning instead leveraged 3D point cloud, a more compact and intuitive representation compared to volumes and image sets for feature learning.

A daunting task in deep learning with 3D point clouds is how to let a neural network consume a point cloud properly because mathematically, a point cloud is a set, and so to define a valid convolution for a point cloud, it is necessary to ensure that the convolution is invariant to the permutation of the point set. 
PointNet~\citep{qi2017pointnet} pioneered the first point cloud convolution with global features by max-pooling per-point features from MLPs. 
Several follow-up works focus on designing convolutions that can learn local features for a point cloud efficiently~\citep{hua2017point,qi2017pointnet++,li2018pointcnn,xu2018spidercnn,wang2018edgeconv,zhang-shellnet-iccv19}.
Interested readers could refer to the survey by Guo et al.~\citep{guo-point-survey-2019} for a comprehensive overview of deep learning techniques for 3D point clouds. 

However, a missing property in the previously mentioned convolution for point clouds is rotation invariance. 
To handle rotations, a common approach is to augment the training data with arbitrary rotations, but a disadvantage of this approach is that generalizing the predictions to unseen rotations is challenging, not to mention that the training time becomes longer due to the increased amount of training data. 
Instead, it is desirable to design a specific convolution with rotation-invariant features. 
In the 2D domain, rotation invariance is an appealing property in feature learning where various methods have been proposed such as learning steerable filters~\citep{weiler2018learning}, performing a log-polar transform of the input~\citep{esteves2018polar} with cylindrical convolutional layers~\citep{kim2020cycnn}, or
implementing transformation invariant pooling operators~\citep{laptev2016ti}. However, these methods are not directly applicable to 3D point clouds due to their difference in both data representation and data dimensionality.

In the 3D domain, rotation invariance has also been specifically built for feature learning of point clouds. 
\citep{rao-spherical-cvpr19} mapped a point cloud to a spherical domain to define a rotation-invariant convolution. However, the learned features are not purely rotation invariant as the discretized sphere by itself is sensitive to global rotations, resulting in a notable performance drop for objects with arbitrary rotations. 
To improve the rotation invariant capacity, \citep{poulenard-spherical-3dv19} proposed to integrate spherical harmonics to a convolution. 
\citep{chen2019clusternet} introduced a hierarchical clustering scheme to encode the relative angles between two-point vectors, and vector norm to keeps rotation invariance.
\citep{zhang-riconv-3dv19} proposed a simple convolution that operates on handcrafted features built from Euclidean distances and angles that are rotation invariant by nature. While consistent results are achieved for arbitrary rotations, only local features are considered which are less descriptive and can cause accuracy degradation. 
Their follow-up work~\citep{zhang2020global} addressed this limitation by building a global context aware convolution based on anchors and Local Reference Frame (LRF) to achieve rotation invariance. \citep{kim2020rotation} learned rotation invariant local descriptors to aggregate local features based on LRF and applies graph convolutional neural networks.
\citep{thomas2020rotation} also relied on LRF and used multiple alignment scheme to gain better results.
\citep{li2021rotation} presented an effective framework to construct both local and global features based on distances, angles and reference points. However, neither LRF nor the global reference are stable enough under noise or outliers, limiting their overall performance. 

A great benefit of having rotation invariance property during feature learning is that it allows \emph{consistent} predictions across training/testing scenarios with or without rotations being applied to the data, and they can generalize robustly to inputs with unseen rotations. 
However, we found that the existing techniques share a common drawback: their performance is inferior to that of a translation-invariant point cloud convolution. 
This is well reflected in the accuracy of the object classification task on ModelNet40 dataset~\citep{wu-3dshapenets-cvpr15}. 
State-of-the-art translation-invariant convolutions such as PointNet~\citep{qi2017pointnet}, PointNet++~\citep{qi2017pointnet++}, PointCNN~\citep{li2018pointcnn}, or ShellNet~\citep{zhang-shellnet-iccv19} report 89\%--93\% of accuracy while techniques with rotation-invariant convolution only report up to 86\%--89\% of accuracy~\citep{zhang-riconv-3dv19,poulenard-spherical-3dv19,zhang2020global,kim2020rotation,li2021rotation}. 
Our work in this paper is dedicated to analyze and reduce this performance gap.

\section{Our Method}
\label{method}

\subsection{Problem Definition}
Our goal is to seek a simple but efficient way to perform convolution on a point set such that the translation and rotation invariance property is preserved.
Mathematically, let $P$ be the point set; we aim for
\begin{align}
    \mathrm{conv}(\pi(P)) = \mathrm{conv}(P)
\end{align}
where $\pi()$ denotes an arbitrary rotation, translation, or point permutation.  
The traditional convolution is translation invariant by definition. 
Convolution for a point cloud is specially designed to achieve the permutation invariance property~\citep{qi2017pointnet}. 
However, techniques that allows point cloud convolution with rotation invariance property has been so far scarce.

An effective solution is to make the input features of $P$ invariant to both translations and rotations, and then design a convolution operator that achieves permutation invariance. 
In this work, we propose to achieve rotation invariance for point cloud convolution with the state-of-the-art performance by directly leveraging rotation invariant features drawn from low-level geometric cues in the Euclidean space. 
For completeness, let us first discuss the design of rotation-invariant features in an early version of this work (RIConv~\citep{zhang-riconv-3dv19}), and then discuss an improved version (\ourconv) with more distinctive features.

\begin{figure}[t!]
\centering
\includegraphics[width=0.8\linewidth]{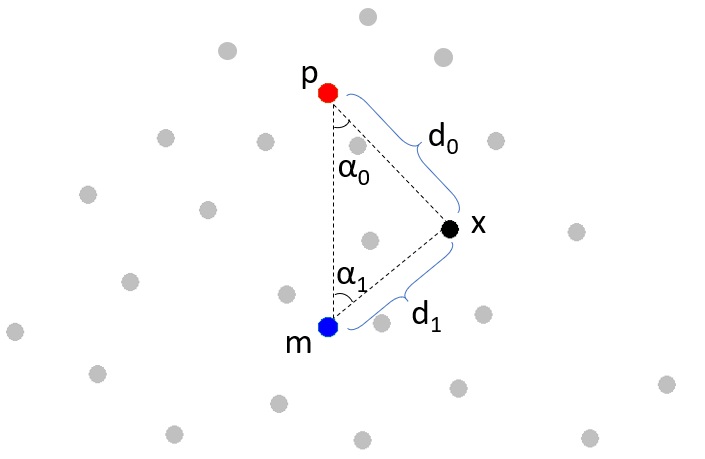}
\caption{Rotation invariant features at a point $x$ used by RIConv~\citep{zhang-riconv-3dv19}. The distances and angles are constructed based on the reference vector $\vec{pm}$, where $p$ is the representative point and $m$ the centroid of the local point set. Such features give good performance in the classification task, but are not as distinctive as features learned from translation-invariant convolutions.}
\label{fig:rif}
\end{figure}

\begin{figure*}[t!]
\centering
\includegraphics[width=\linewidth]{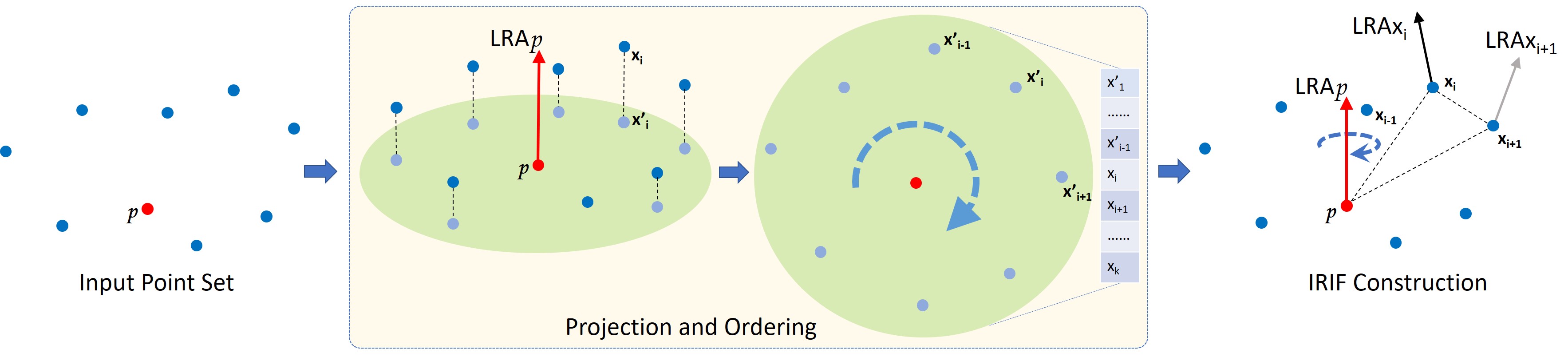}
\parbox[b]{1\textwidth}{\relax
		\quad \qquad \quad \quad \qquad (a) \quad \qquad \qquad \qquad  \quad \qquad \qquad \qquad \qquad (b) \qquad \qquad \qquad \quad \qquad \qquad \qquad \qquad\qquad\qquad(c)}
\caption{Informative rotation invariant features (IRIF). The local reference axis (LRA) indicates the local orientation. For a point set with the red point $p$ as reference and grey points as its neighbors (a), clockwise ordering is imposed by projection of the points onto the local tangent disk (b). At each neighbor point, we compute the informative rotation invariant features from the relations between $p$ and its neighbors as well as the relations between the neighbors (c).}
\label{fig:irif}
\end{figure*}

\subsection{Rotation Invariant Local Features} 
\label{sec:irif}
The design of rotation-invariant features used in RIConv can be explained as in Figure~\ref{fig:rif}. 
Given a reference point $p$ (red), $K$ nearest neighbors are determined to construct a local point set. 
The centroid of the point set is denoted as $m$ (blue). 
We use vector $\vv{pm}$ as a reference to extract translation and rotation invariant features for all points in the local point set. 

Particularly, for a point $x$ in this set, its features are defined as
\begin{equation}
	\mathrm{RIF}(x) = [d_{0}, d_{1}, \alpha_{0}, \alpha_{1}]\,.
\end{equation}
Here, $d_{0}$ and $d_{1}$ represent the distances from $x$ to $p$ and to $m$, respectively. 
$\alpha_{0}$ and $\alpha_{1}$ represent the angles from $x$ towards $p$ and $m$, as shown in Figure~\ref{fig:rif}. 
Since such low-level geometric features are invariant under rigid transformations, they are very well suited for our need to make a translation invariant convolution with rotation invariance property. 
Note that the reference vector $\vv{pm}$ can also serve as a local orientation indicator and we can use it to build a local coordinate system for convolution.
Such rotation invariant features have been used in RIConv~\citep{zhang-riconv-3dv19} that achieves good classification results on the ModelNet40 dataset~\citep{wu-3dshapenets-cvpr15}. 



A caveat from the feature extraction scheme above is the stability of the vector $\vv{pm}$. 
When the centroid $m$ changes, it can cause $\vv{pm}$ to be unstable. 
In this work, we introduce local reference axis (LRA), a more stable reference vector based on the theory of local reference frame (LRF) for rotation-invariant shape descriptors. 
Similar to vector $\vv{pm}$, LRA can be used for extracting rotation-invariant features, and for indicating the orientation of the local neighborhood to define the convolution subsequently.

\paragraph{Local Reference Axis}

Given point $p$ and its neighbors $x_i\in \Omega_p$. 
A local reference axis (LRA) at $p$ is defined as the eigenvector corresponding to the smallest eigenvalue of the covariance matrix:
\begin{equation}
\Sigma = \sum_{i=1}^{N_{sub}} w_i (x_i - p) (x_i - p)^\top,
\end{equation}
where $N_{sub}$ is the number of points in the local region and $x_i\in \Omega_p$, and
\begin{equation}
w_i = \frac{m - \| x_i - p \|}{\sum_{i=1}^{N} m - \| x_i - p \|},
\end{equation}
where $m = \max_{i=1..N_{sub}}(\| x_i - p \|)$. 
Intuitively, this weight allows nearby points of $p$ to have large contributions to the covariance matrix, and thus greatly affect the LRA. 
Points further away from $p$ however can contribute globally to the robustness of the LRA. 
Such weighted LRA construction is a fundamental step in 3D hand-crafted features~\citep{tombari2010unique}, which can be easily integrated into our proposed convolution.

Note that the construction of LRA is very similar to that of local coordinate frame (LRF) used in traditional hand-crafted shape descriptor~\citep{tombari2010unique}. 
It can be regarded that LRA is the most stable part of LRF (see Section~\ref{sec:analysis} for the comparison experiment); in most cases (e.g., locally flat surfaces), LRA is highly similar to the normal vector of the surface. 
Empirically, we found that features learned with LRA performs as well as those learned with normal vectors. 
Note that compared to LRF, we do not make use of the two axes tangential to the surface because they could be ambiguous and unstable. 

\begin{figure}[tb]
\centering
\includegraphics[width=0.8\linewidth]{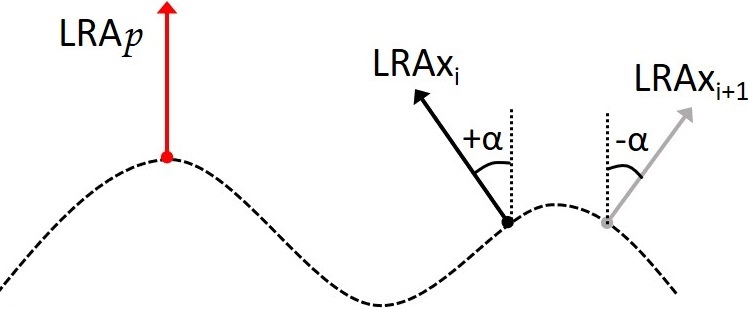}
\caption{Illustration of signed angles. Although the absolute angle values  of $LRA_{x_{i+1}}$ and $LRA_{x_{i}}$ relative to $LRA_p$ are the same, their directions are opposite. So the signs are used to indicate the directions.}
\label{fig:signed_angle}
\end{figure}

\paragraph{Informative Rotation Invariant Features.}
Given the definition of LRA, we propose more powerful rotation-invariant features as follows. 
Given a reference point $p$, recall that RIConv~\citep{zhang-riconv-3dv19} only considers the relation between $p$ and its neighbors, measuring Euclidean distances and angles as rotation-invariant features. 
We propose to additionally consider the distances and angles among the neighbors themselves. 
An illustration of our proposed features is shown in Figure~\ref{fig:irif}. 
In this design, the features are kept rotation invariant by definition, and so the framework of RIConv~\citep{zhang-riconv-3dv19} can work as is.
We name such features \emph{Informative} Rotation Invariant Features (IRIF). IRIF transforms each of the neighbor point $x_i$ into a tuple of seven attributes:
\begin{equation}
\label{eq:5}
	\mathrm{IRIF}(x_i) = [d, \varphi, \alpha_{0}, \alpha_{1},  \alpha_{2}, \beta_{0}, \beta_{1}, \beta_{2}]\,
\end{equation}
where  $d$, $\alpha_{0}$, $\alpha_{1}$ and $\alpha_{2}$ measure the relationship between neighbor point $x_i$ and the reference point $p$ (radial direction):
\begin{align}
d &= \| x_i - p \|,\\
\alpha_{0} &= \angle\left( LRA_{x_{i}}, \vv{x_{i}p} \right), \nonumber\\
\alpha_{1} &= \angle\left( LRA_{p}, \vv{x_{i}p} \right), \nonumber\\
\alpha_{2} &= S_{a} \cdot \angle\left( LRA_{x_{i}}, LRA_{p} \right). \nonumber
\end{align}
and $\varphi$, $\beta_{0}$, $\beta_{1}$, $\beta_{2}$ encode the relationship between $x_i$ and its adjacent neighbor $x_{i+1}$ (clockwise direction):
\begin{align} 
\label{eq:7}
\varphi &= \angle\left( \vv{x_{i+1}p} , \vv{x_{i}p} \right), \\
\beta_{0} &= \angle\left( LRA_{x_{i}}, \vv{x_{i}x_{i+1}} \right), \nonumber\\
\beta_{1} &= \angle\left( LRA_{x_{i+1}}, \vv{x_{i}x_{i+1}} \right), \nonumber\\
\beta_{2} &= S_{b} \cdot \angle\left( LRA_{x_{i}}, LRA_{x_{i+1}} \right). \nonumber
\end{align}
Here, $\angle$ is computed as the arccos of the normalized vectors. Since arccos returns values in $[0, \pi]$, it has a signed ambiguity as shown Figure~\ref{fig:signed_angle}. Although the absolute angle values are the same, their directions with regard to $LRA_{p}$ are totally different. To differentiate this, in Equation~\ref{eq:5}, we propose signed angle to encode both of the angle and direction information between two vectors. We define $S_{a}$ and $S_{b}$ to encode the directions as
\begin{align}
   S_{a} &=
   \begin{cases}
      +1, & \text{if}\ \alpha_{0} \leq \alpha_{1} \\
      -1, & \text{otherwise}
    \end{cases},\quad  
    S_{b} &=
    \begin{cases}
      +1, & \text{if}\ \beta_{0} \leq \beta_{1} \\
      -1, & \text{otherwise}
    \end{cases}. \nonumber
\end{align}
\paragraph{Uniqueness of IRIF.}
During the IRIF construction, it is expected that each point is converted to an unique position in the feature space. The attributes $d$, $\alpha_{0}$, $\varphi$ in Equation~\ref{eq:5} correspond to the radial distance, polar angle, azimuthal angle respectively which uniquely define a 3D point in the local system. Other attributes are used to encode the second order properties making our feature more informative. Note that, for each neighbor point $x_{i}$ the attribute $\varphi$ is the angle between its clockwise adjacent neighbor $x_{i+1}$, and its angle with other neighbor points can be obtained by chain rule. So, each neighbor point has an unique position in the local spherical system theoretically. However, there is a special case that all neighbor points are uniformly distributed along azimuthal direction with exactly same radial distances and same polar angles (e.g., a sphere). In this case, IRIF is no longer unique, but IRIF would still function as a feature with reduced descriptiveness.

Additionally, for the tasks targeting whole objects like classification and retrieval, global uniqueness is also important. This is achieved by enlarging the neighborhood size to include all the interest points. For example, the input of last convolution layer usually contains less number of points with higher dimensions. We can take all the input points into in the last layer to achieve global uniqueness.

\begin{figure*}[t]
\centering
\includegraphics[width=0.95\textwidth]{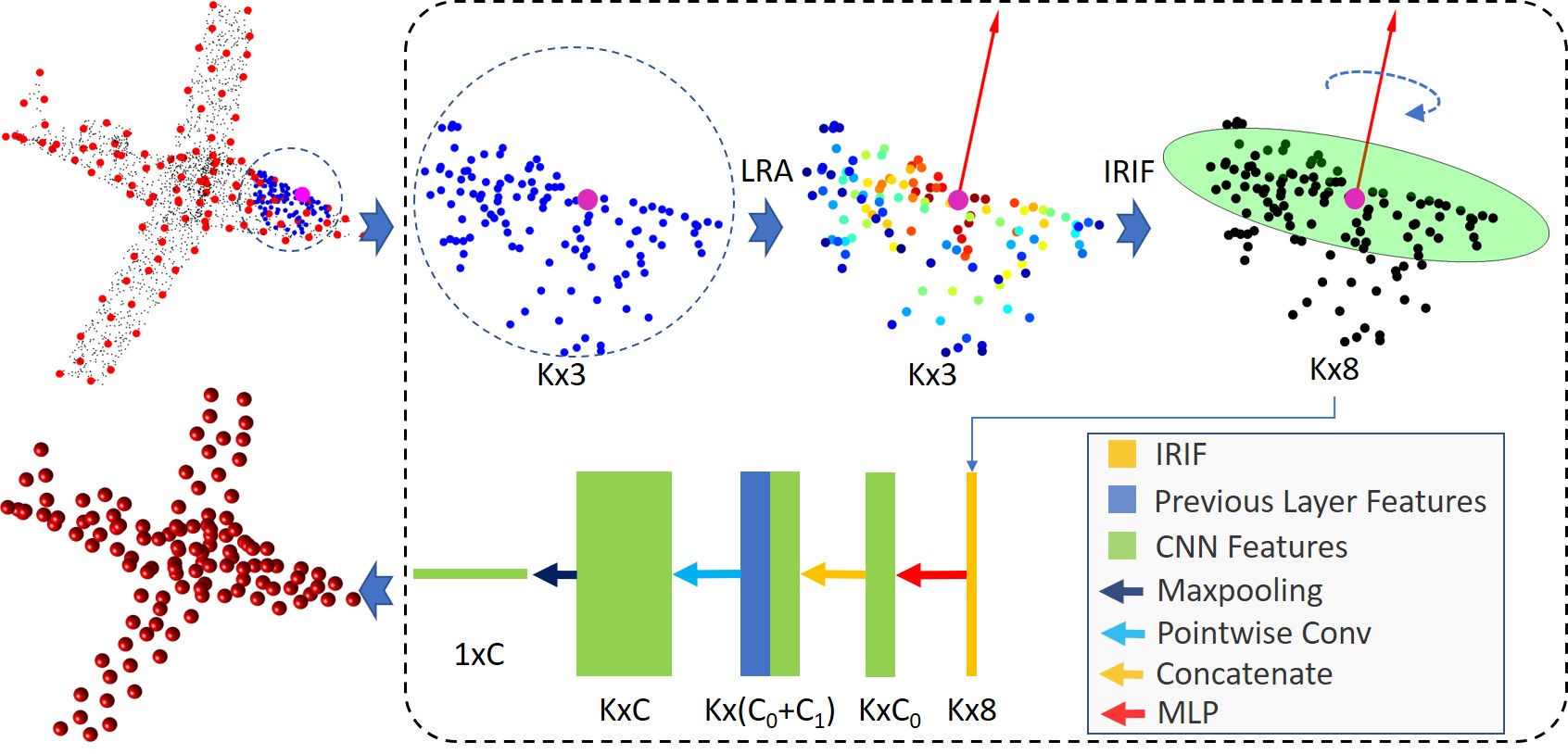}
\caption{\ourconv operator. For an input point cloud, representative points (red dots) are sampled via farthest point sampling. For a reference point $p$ (pink), K nearest neighbors are queried to yield a local point set. Then, we compute the informative rotation invariant features (Section~\ref{sec:irif}), which is lifted to a high-dimensional space by a shared multi-layer perceptron (MLP). Concatenated with previous layer features (if any), the features of these local points are further passed to a pointwise convolution, which are finally summarized by maxpooling (Section~\ref{sec:conv}). The convolution is applied to all representative points, resulting in output features at such features (denoted as thicker red points).}
\label{fig:riconvpp}
\end{figure*}

\begin{algorithm*}[t]
	\small
	\caption{\ourconv operator.}
	\label{alg:conv}
	\begin{algorithmic}[1]
		\Require
		Reference point $p$, point set $\Omega$, point features $\mathbf{f}_{prev}$ from previous layer (if any), local reference axes $LRA$
		\Ensure
		Convoluted features $\mathbf{f}$
		\State $\mathbf{f} \leftarrow \left\{ \mathrm{IRIF}(x_i) : \forall x_i \in \Omega \right\}$ \hfill * Construct informative rotation invariant features with LRA axes (Section~\ref{sec:irif})
		\State $\mathbf{f} \leftarrow \mathrm{MLP}(\mathbf{f})$; \hfill * Lift each feature to a high-dimensional feature
		\State $\mathbf{f}_{in} \leftarrow [\mathbf{f}_{prev}, \mathbf{f}]$ \hfill * Concatenate the features from the local and the previous layer (if any)
		\State $\mathbf{f}_{out} \leftarrow \mathrm{conv}(\mathbf{f}_{in})$ \hfill *  1D convolution with proper ordering\\
		\Return $\mathrm{maxpool}(\mathbf{f}_{out})$ \hfill * Maxpool features and return
	\end{algorithmic}
\end{algorithm*}

\subsection{Rotation-Invariant Convolution}
\label{sec:conv}

From the recipes of informative rotation invariant features and local reference axis, we are now ready to define our convolution. The main steps are detailed in Figure~\ref{fig:riconvpp}. 

Particularly, we start by sampling a set of representative points through farthest point sampling strategy which can generate uniformly distributed points. 
From each of which we perform a set of K-nearest neighbors to obtain local point sets. 
Let us consider a local point set $\Omega = \{x_i\}$ where $x_i$ represents 3D coordinates of the point $i$. 
We define the convolution to learn the features of $\Omega$ as
\begin{align} 
\mathbf{f}(\Omega) = \sigma( \mathcal{A} ( \{ \mathcal{T}(\mathbf{f}_{x_i}) : \forall i \} ) )
\end{align}
This formula indicates that features of each point in the point set are first transformed by $\mathcal{T}$ before being aggregated by the aggregation function $\mathcal{A}$ and passed to an activation function $\sigma$. 
We set the input features to our informative rotation-invariant features $\mathbf{f}_{x_i} = \mathrm{IRIF}(x_i)$.
We define the transformation function as
\begin{align}
\mathcal{T}(\mathbf{f}_{x_i}) = \mathbf{w}_i \cdot \mathbf{f}_{x_i} = \mathbf{f}'_{x_i}
\end{align}
where $\cdot$ indicates the element-wise product, and $\mathbf{w}_i$ is the weight parameter to be learned by the network. 
Our transformation function is similar to PointNet~\citep{qi2017pointnet}, but applied locally.

\begin{figure*}[t]
\centering
\includegraphics[width=0.96\textwidth]{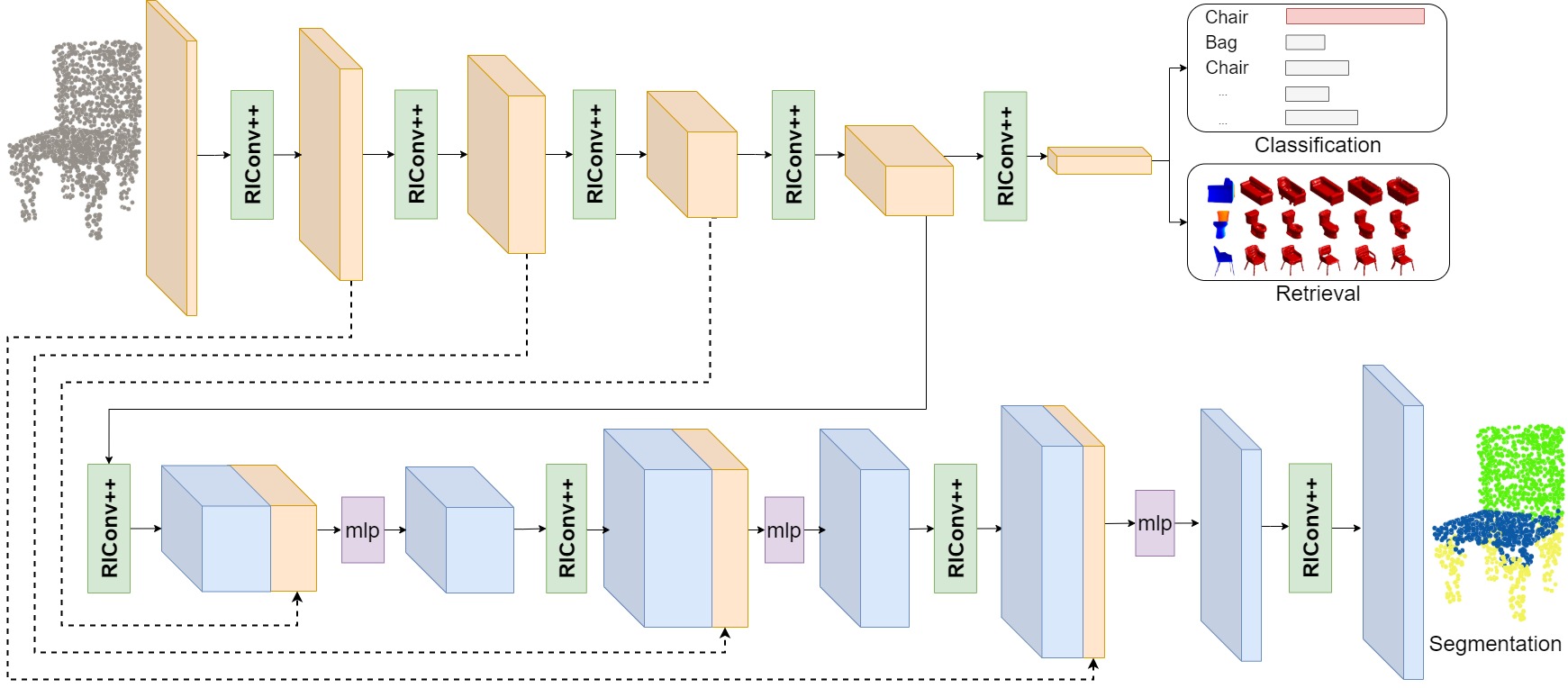}
\caption{Our network architecture comprises five convolution layers to extract point cloud features before fully connected layers for object classification and shape retrieval task. 
We add a decoder with skip connections for segmentation task.}
\label{fig:rinet}
\end{figure*}

A popular choice of the aggregation function $\mathcal{A}$ is maxpooling, which supports permutation invariance in the orders of the input point features~\citep{qi2017pointnet}.
Our aggregation function differs in that it includes a 1D convolution kernel and an ordering function to maintain rotation invariance. This has been used in our previous work~\citep{zhang-riconv-3dv19,zhang-shellnet-iccv19}:
\begin{equation}
     \mathcal{A}(\left\{ \mathbf{f}'_{x_i} \right\}) = \mathrm{maxpool} \left( \mathrm{K} \star \mathrm{order}(\left\{ \mathbf{f}'_{x_i} \right\} ) \right) 
\end{equation}
where $\mathrm{order}$ is a function that sorts the points in a clockwise order based on projecting $x_i$ to the local tangent disk, and $K$ is the 1D convolution kernel. 
To obtain the ordering, we select one neighbor point as starting point (e.g. $x_{0}$), and set $\vv{x_{0}p}$ as reference on the projected disk. Then, we compute the angles between $\vv{x_{i}p}$ and $\vv{x_{0}p}$ the ordering is determined by sorting the angles from 0 to 360 degrees.
In our implementation, we simply select $x_{0}$ by farthest point in the local neighborhood from the reference point $p$.
A benefit of using the farthest point is that the distance between the farthest point and the reference point is less likely to be zero, and so the IRIF features can be validly computed. 
Note that when the kernel size is $1$, point ordering is only used for feature encoding (Equation~\ref{eq:7}) and not necessary for convolution, and the maxpooling makes the resulting features invariant to point permutation. 
The detailed steps to perform convolution is shown in the Algorithm~\ref{alg:conv}.

Compared to RIConv~\citep{zhang-riconv-3dv19} and GCAConv~\citep{zhang2020global}, our convolution is simpler as it does not require binning. In fact, one of the effects of binning is to fix the instability of the $\vv{pm}$ vector. In our case, as we find that the LRA is sufficiently stable, we simply apply pointwise convolution as described. 

In addition, traditional convolutional neural networks often allow downsampling/upsampling to manipulate the spatial resolution of the input. We build this strategy into our convolution by simply treating the sampled point set as the downsampling/upsampling points.

\subsection{Network Architecture}
\label{sec:network}

\begin{table}[t]
    \caption{The details of our neural network. Refer to Figure~\ref{fig:rinet} for an illustration of the network architecture, and Algorithm~\ref{alg:conv} for steps in \ourconv operator. Here, K is the number of categories, and N is the number of input points.}
    \label{tab:network}
    \centering
    \begin{tabular}{l | c} 
        \noalign{\smallskip}\hline\noalign{\smallskip}
        Module & Output shape\\
        \noalign{\smallskip}\hline\noalign{\smallskip}
        \textbf{\ourconv} & \\
        Input tensor  & in\_dims $\times$ in\_points \\
        RIConv++ operator & out\_dims $\times$ out\_points \\
        BatchNorm & out\_dims $\times$ out\_points \\
        ReLU &  out\_dims $\times$ out\_points \\
        \\
        \textbf{Classification / Retrieval} & \\
        Input tensor & $3 \times N$ \\
        \ourconv & $32 \times 1024$ \\
        \ourconv & $64 \times 512$ \\
        \ourconv & $128 \times 256$ \\
        \ourconv & $256 \times 128$\\
        \ourconv & $512 \times 1$\\
        Fully connected & $512 \times 1$ \\
        Fully connected & $256 \times 1$ \\ 
        Softmax & $K \times 1$ \\
        \\
        \textbf{Segmentation} & \\
        Input tensor & $3 \times N$ \\
        \ourconv & $64 \times 512$ \\
        \ourconv & $128 \times 256$ \\
        \ourconv & $256 \times 128$ \\
        \ourconv & $512 \times 64$
        \vspace{0.05in}\\
        \ourconv & $512 \times 128$ \\
        Skip connection & $768 \times 128$ \\
        MLP & $512 \times 128$ \\
        \ourconv & $512 \times 256$ \\
        Skip connection & $640 \times 256$ \\
        MLP & $256 \times 256$ \\
        \ourconv & $256 \times 512$ \\
        Skip connection & $320 \times 512$ \\
        MLP & $128 \times 512$ \\
        \ourconv & $K \times N$ \\
        \noalign{\smallskip}\hline\noalign{\smallskip}
    \end{tabular}
\end{table}

We use the proposed convolution to design neural networks for object classification, object part segmentation, semantic segmentation, and shape retrieval, respectively. The architecture is shown in Figure~\ref{fig:rinet} and Table~\ref{tab:network}. Our classification network has a standard architecture and uses five consecutive layers of convolution (with point downsampling) followed by fully connected layers to output the probability map. As our convolution operator is already designed to handle arbitrary rotation and point orders, we can simply place each convolution one after another. By default, each convolution is followed by a batch normalization and an ReLU activation. 

The segmentation network follows an encoder-decoder architecture with skip connections similar to U-Net~\citep{ron2015unet}. 
We use an MLP after a skip connection to unify and transform the combined features to have a valid size before applying a deconvolution. 

The classification network acts as the encoder, yielding the features in the latent space that can be subsequently decoded into part labels. 
Unless otherwise mentioned, we use 1024 points for classification/retrieval, 2048 points for part segmentation, and 4096 points for semantic segmentation, respectively. 

Our deconvolution is detailed as follows. We define deconvolution in the same way as \ourconv. The difference here is that our convolution outputs to a point subset with more feature channels while deconvolution outputs to a point set with more points compared to the input with fewer feature channels.
Particularly, suppose that deconvolution begins with the set of $N_l$ points at layer $l$.  
The RIConv++ operator is applied and the features of $N_l$ points are upsampled to a set of $N_{l+1}$ points at the next layer $l+1$. 
The deconvolution is repeatedly applied until the
point cloud reaches the original number of points $N$.
Note that as we have the point subsets during downsampling in the encoder part, we do not need to generate points in upsampling, but just need to reuse the subsets and propagate the features by interpolation.

\paragraph{Neighborhood Size.} 
Our framework is able to integrate both local and global features by simply adjusting the neighborhood size. For classification and retrieval tasks, the global information is more important, so we set the the nearest neighbor size as $8$, $16$, $32$, $64$, and $128$ respectively for the five layers of convolutions in the encoder to extract features from local to global. 
For segmentation, we use the same setting for the encoder but carry the features from the 4th layer to the decoder to focus more on the local features and set the neighborhood size as $8$, $16$, $32$, and $32$ for the decoder layers.

\section{Experiments}
\label{sec:exp}

\begin{table*}[t]
\centering
\caption{Comparisons of the classification accuracy (\%) on the ModelNet40 dataset. On average, our method has the best accuracy and lowest accuracy deviation in all cases including with and without normals. Here, 'nor' means 'normal'.}
\label{tab:classification_modelnet40}
\begin{tabular}{l|l|llr|ccc|cc}
\noalign{\smallskip}\hline\noalign{\smallskip}
& Method  & Format & Input Size & Params.& z/z$\uparrow$ & SO3/SO3$\uparrow$ &z/SO3$\uparrow$ & Std.$\downarrow$  \\
\noalign{\smallskip}\hline\noalign{\smallskip}

\multirow{7}{*}{\rotatebox[origin=c]{90}{Traditional}} & VoxNet~\citep{maturana2015voxnet} & voxel & $30^{3}$ &0.90M & 83.0 & 87.3 & - & 3.0 \\
& SubVolSup~\citep{qi2016volumetric}& voxel & $30^{3}$ &17.00M  & 88.5 & 82.7 & 36.6 & 28.4 \\ 
		& MVCNN 80x~\citep{su2015multi}  & view & $80\times 224^{2}$ & 99.00M & 90.2 & 86.0 & 81.5 & 4.3 \\
		& PointNet~\citep{qi2017pointnet}  & xyz & $1024\times 3$ & 3.50M & 87.0 & 80.3 & 21.6 & 41.0 \\ 
		& PointCNN~\citep{li2018pointcnn}  & xyz & $1024\times 3$ & 0.60M & 91.3 & 84.5 & 41.2 & 27.2 \\
		& PointNet++~\citep{qi2017pointnet++}  & xyz + nor & $1024\times 6$ & 1.40M & 89.3 & 85.0 & 28.6 & 33.8\\
		& DGCNN~\citep{wang2018edgeconv}  & xyz & $1024\times 3$ & 1.84M & 92.2 & 81.1 & 20.6 & 38.5 \\
		& RS-CNN~\citep{liu2019relation}  & xyz & $1024\times 3$ & 1.41M & 90.3 & 82.6 & 48.7 & 22.1\\
\noalign{\smallskip}\hline\noalign{\smallskip}
\multirow{12}{*}{\rotatebox[origin=c]{90}{Rotation-invariant}} & Spherical CNN~\citep{esteves2018learning}  & voxel & $2\times 64^{2}$ & 0.50M & 88.9 & 86.9 & 78.6 & 5.5\\
& RIConv ~\citep{zhang-riconv-3dv19} & xyz & 1024 $\times 3$ & 0.70M & 86.5 & 86.4 & 86.4 & 0.1\\
& SPHNet ~\citep{poulenard-spherical-3dv19} & xyz & 1024 $\times 3$ &  2.90M & 87.0 & 87.6 & 86.6 & 0.5 \\
& SFCNN~\citep{rao-spherical-cvpr19} & xyz & 1024 $\times 3$ &  - & \textbf{91.4} & 90.1 & 84.8 & 3.5 \\
& ClusterNet~\citep{chen2019clusternet} & xyz & 1024 $\times 3$ &  1.40M & 87.1 & 87.1 & 87.1  & \textbf{0.0} \\
& GCAConv~\citep{zhang2020global} & xyz & 1024 $\times 3$ & 0.41M & 89.0 & 89.2 & 89.1 & \textbf{0.0}\\
& RI-GCN~\citep{kim2020rotation} & xyz & 1024 $\times 3$ & 4.38M & 89.5 & 89.5 & 89.5 & \textbf{0.0}\\
& RIF~\citep{li2021rotation} & xyz & 1024 $\times 3$ & - & 89.4 & 89.3 & 89.4 & \textbf{0.0}\\
& Ours  & xyz & 1024 $\times 3$ & \textbf{0.42M} & 91.2 & \textbf{91.2} & \textbf{91.2}  & \textbf{0.0}\\
\noalign{\smallskip}\cline{2-9}\noalign{\smallskip}
& RI-GCN~\citep{kim2020rotation} & xyz + nor & 1024 $\times 6$ & 4.38M & 91.0 & 91.0 & 91.0 & \textbf{0.0}\\
& Ours  & xyz + nor & 1024 $\times 6$ & \textbf{0.42M} & \textbf{91.3} & \textbf{91.3} & \textbf{91.3} & \textbf{0.0}\\
\noalign{\smallskip}\hline
\end{tabular}
\end{table*}

We report our evaluation results in this section. We implemented our network in PyTorch, and use a batch size of $16$ for all the tasks in training. 
The optimization is done with an Adam optimizer. The initial learning rate is set to 0.001. Our training is executed on a computer with an Intel(R) Core(TM) i7-10700K CPU equipped with a NVIDIA GTX 2080ti GPU. 

We evaluate our method with object classification, shape retrieval, object part segmentation, and semantic segmentation. 
For object classification and retrieval, we train for $200$ epochs, and the network usually converges within $150$ epochs. 
For object part segmentation, we train for $200$ epochs, and the network usually converges within $150$ epochs. For semantic segmentation, we train for 40 epochs.
It takes about 2 hours for the training to converge for classification, about 15 hours for part segmentation, and about 40 hours for semantic segmentation. 

Following \citet{esteves2018learning}, we perform experiments in three cases: (1) training and testing with data augmented with rotation about gravity axis (z/z), (2) training and testing with data augmented with arbitrary SO3 rotations (SO3/SO3), and (3) training with data by z-rotations and testing with data by SO3 rotations (z/SO3). 
The first case is commonly used for evaluating translation-invariant point cloud learning methods, and the last two cases are for evaluating rotation invariance. 
In general, convolution with rotation invariance is expected to work well for case (3) even though the network is not trained with data augmented with SO3 rotations.

In general, our result demonstrates the effectiveness of the rotation invariant convolution we proposed. 
Our networks yield very \emph{consistent} results despite that our networks are trained with a limited set of rotated point clouds and tested with arbitrary rotations. 
To the best of our knowledge, there is no previous work for point cloud learning that can achieve the same level of accuracy with the same level of consistency despite that some methods~\citep{esteves2018learning,rao-spherical-cvpr19} demonstrated good performance when trained with a particular set of rotations. 
We detail our evaluations below.

\subsection{Object Classification}
\label{sec:classification}

The classification task is trained on the ModelNet40 variant of the ModelNet dataset~\citep{wu20153d}. ModelNet40 contains CAD models from 40 categories such as airplane, car, bottle, dresser, etc. By following~\citet{qi2017pointnet++}, we use the preprocessed $9,843$ models for training and $2,468$ models for testing. The input point cloud size is 1024, with each point has the attributes $(x, y, z, nx, ny, nz)$ which are 3D coordinates and 3D normals in the Euclidean space. 

We use the encoder layers of Figure~\ref{fig:rinet} which outputs one feature vector to train the classifier. Particularly, our network outputs one feature vector of length $512$ to the classifier, which is then passed through an MLP implemented by fully connected layers, resulting in $128 \times 40$ category predictions.

We use two criteria for evaluation: accuracy and accuracy standard deviation. Accuracy is a common metric to measure the performance of the classification task. In addition, accuracy deviation measures the consistency of the accuracy scores in three tested cases. In general, it is expected that methods that are rotation invariant should be insusceptible to the rotation used in the training and testing data and therefore has a low deviation in accuracy.
 
The evaluation results are shown in Table~\ref{tab:classification_modelnet40}.
As can be seen, our method achieves the state-of-the-art performance in all cases. 
More importantly, our method has almost zero accuracy deviation. 
Non-rotation invariant point cloud learning methods exhibit large accuracy deviations especially in the extreme z/SO3 case. 
This case is exceptionally hard for methods that rely on data augmentation to handle rotations~\citep{qi2017pointnet,qi2017pointnet++}. 
In our observation, such techniques are only able to generalize within the type of rotation they are trained with, and generally fail in the z/SO3 test. 
This applies to both voxel-based and point-based learning techniques. 
By contrast, our method has almost no performance difference in three test cases, which confirms the robustness of the rotation invariant geometric cues in our convolution. 
The success of our method is attributed to two factors: the informative rotation-invariant features (Section~\ref{sec:irif}), and the use of large neighborhood in the last layer of the network that naturally widens the perceptive field and captures global features. 

Table~\ref{tab:classification_modelnet40} also includes a comparison on the use of normal vectors for feature learning as follows. 
For our method with input format $xyz$, we use the LRA as the reference axis, while with input format $xyz+nor$, we use normal vectors as the reference axis. 
This is different from PointNet++ and RI-GCN which treat normals as extra features.
In both cases, our method has almost similar performance difference ($0.1\%$ accuracy difference) which means that our LRA is as descriptive as normal vectors. Our method also outperforms both PointNet++ and RI-GCN in both cases. 
Note that we achieve this performance without voting, a test-time augmentation scheme to boost the classification result used by RI-GCN. When voting is disabled, RI-GCN has 1\% accuracy drop. Voting also slows down the inference in real applications.

\paragraph{Network Parameters.} The capability to handle rotation invariance also has a great effect on the number of network parameters. For networks that rely on data augmentation to handle rotations, it requires more parameters to `memorize' the rotations. Networks that are designed to be rotation invariant, such as spherical CNN~\citep{esteves2018learning} and ours, have very compact representations. 
In terms of the number of trainable parameters, our network has 0.4 millions (0.4M) of trainable parameters, which is the most compact network in our evaluations. Among the tested methods, only spherical CNN~\citep{esteves2018learning} (0.5M) and PointCNN (0.6M) have similar compactness. Our network has $9\times$ less parameters than PointNet (3.5M), more than $3\times$ less than PointNet++ (1.4M). The good balance between trainable parameters, accuracy and accuracy deviations makes our method more robust for practical use.

\subsection{An Analysis of Rotation-Invariant Features}
\label{sec:analysis}

We compare the performance with RIConv~\citep{zhang-riconv-3dv19} and GCAConv~\citep{zhang2020global} using same number of layers and same number of representative points in each layer. 
Here, we set the total number of layers as 3 with 512, 128, 32 representative points and 64, 32, 16 nearest neighbors, respectively. 
The results are shown in Table~\ref{tab:compare_rifeat}.
In general, we observe that our informative rotation invariant features (IRIF) is more accurate than original features used in RIConv~\citep{zhang-riconv-3dv19} and GCAConv~\citep{zhang2020global}.

\begin{table}[t]
\caption{Performance comparisons with our previous work, RIConv ~\citep{zhang-riconv-3dv19} and GCAConv ~\citep{zhang2020global} (\%) on the ModelNet40 dataset with the same representative points and same number of layers. It shows the effectiveness of our proposed features.}
\label{tab:compare_rifeat}
	\centering
	\footnotesize
	\begin{tabular}{l|cccc}
		\noalign{\smallskip}\hline\noalign{\smallskip}
		Method   & Core Features & OA \\
		\noalign{\smallskip}\hline\noalign{\smallskip}
		RIConv ~\citep{zhang-riconv-3dv19} & RIF  &86.5  \\
		GCAConv ~\citep{zhang2020global} & LRF Transform  &89.0\\	
		\noalign{\smallskip}\hline\noalign{\smallskip}
		Ours (w/o normal)   & IRIF  & \textbf{89.8} \\
		Ours (w/ normal)    & IRIF  & \textbf{90.3} \\
		\noalign{\smallskip}\hline
	\end{tabular}
\end{table}

\begin{figure}[t]
	\centering
	\includegraphics[width=\linewidth]{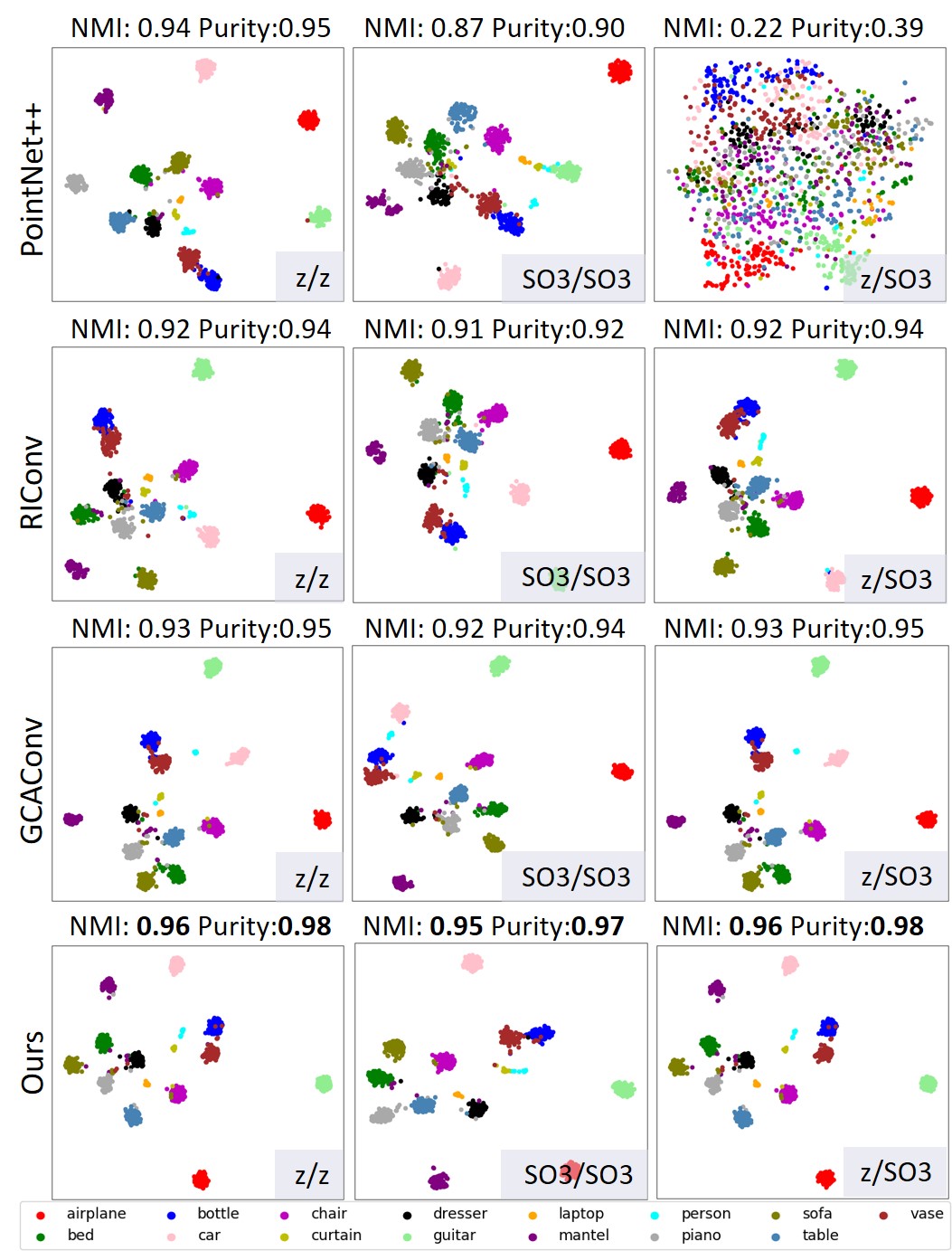}
	\caption{t-SNE comparisons of the latent features for PointNet++~\citep{qi2017pointnet++}, RIConv~\citep{zhang-riconv-3dv19}, GCAConv~\citep{zhang2020global} and our method under three different rotation settings. The clusters in the t-SNEs show that to make good decisions in object classification, it is desirable to have the cluster boundaries as separated as possible.}
	\label{fig:tsne}
\end{figure}

We further visualize the latent space learned by the neural networks using t-SNE~\citep{maaten-tsne-2008}. The results are shown in Figure~\ref{fig:tsne}. 
We measure the clustering quality by two metrics, the normalized mutual information (NMI) and purity~\citep{manning2008introduction}.
We demonstrate three scenarios for object classification: z/z, SO3/SO3, and z/SO3. 
As can be seen, latent space learned by rotation-invariant convolution such as RIConv~\citep{zhang-riconv-3dv19} does not exhibit good discrimination among classes. 
The main difference between such convolution and traditional point cloud convolution is that it no longer works with point coordinates at start. 
In the case of RIConv, the points are transformed into Euclidean-based features including distances and angles, which are not as unique as point coordinates since many points can share the same distance and angles. 
This is well reflected in the t-SNE in the first column (z/z) in Figure~\ref{fig:tsne}. 
PointNet++~\citep{qi2017pointnet++} has a good separation among the clusters while RIConv~\citep{zhang-riconv-3dv19} has more condensed clusters in the center, resulting in more ambiguities during classification. The clusters by GCAConv~\citep{zhang2020global} and our method are more similar to PointNet++, which explains their similar performance.

Similarly, in the second column (SO3/SO3), all methods have similar clustering, which explains their similar performance in the classification (see more quantitative comparisons in Table~\ref{tab:classification_modelnet40}). 
Finally, the third column (z/SO3) highlights the strength of rotation-invariant convolutions as they can still maintain consistent predictions and generalize well to unseen conditions. 
In this case, the t-SNEs show that PointNet++ cannot generalize effectively. 

\subsection{Robustness to Noise}
\label{sec:noise}
\begin{figure}[t]
\centering
\includegraphics[width=0.9\linewidth]{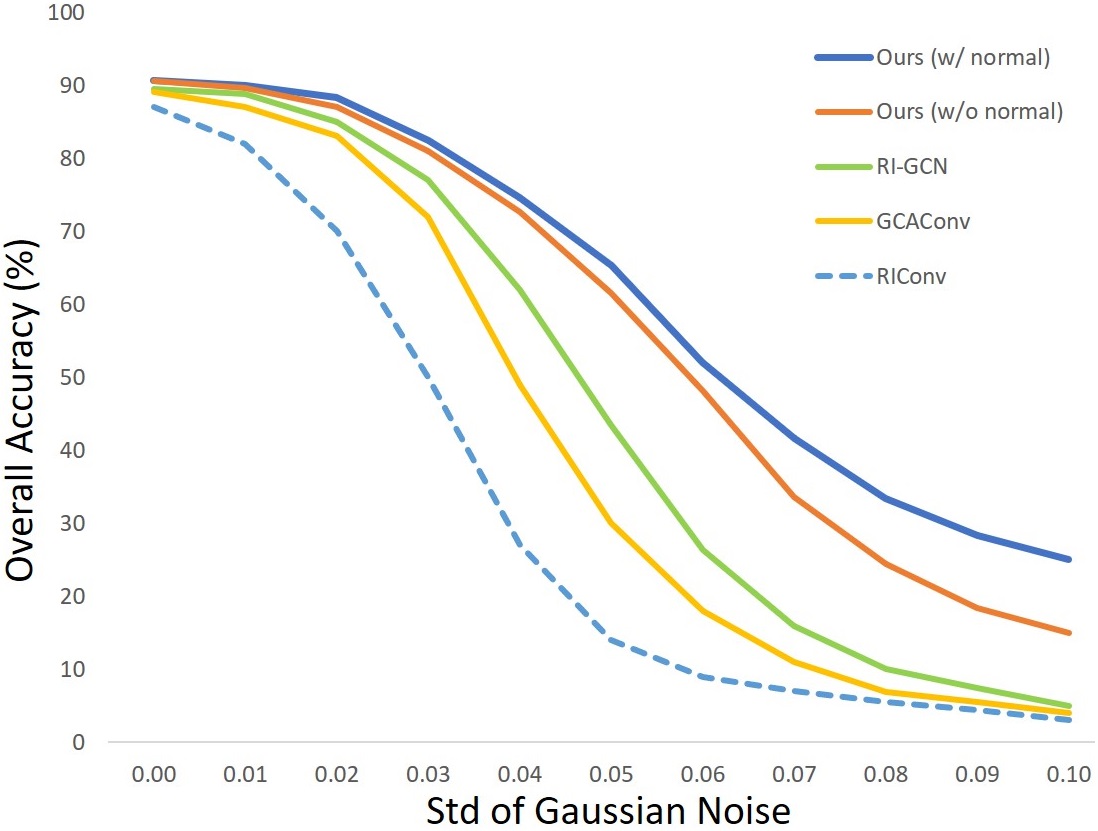}
\caption{Performance of our method in the presence of additive Gaussian noise on the ModelNet40 dataset. Our method outperforms all existing rotation-invariant convolutions.}
\label{fig:noise}
\end{figure}
In real applications, the point cloud data produced from scanners usually contains noise. 
Here we conduct an experiment to verify the robustness of existing rotation-invariant convolutions to noise. 
We sample and add noise from zero-mean Gaussian distributions $N(0,\sigma^{2})$ to the data, and compare the classification performance of our method with RIConv~\citep{zhang-riconv-3dv19}, GCAConv~\citep{zhang2020global} and RI-GCN~\citep{kim2020rotation}. 
The noise level can be controlled by adjusting $\sigma^2$. 
In Figure~\ref{fig:noise}, we plot the accuracy against different noise level $\sigma$. 
It can be seen that our method outperforms all remaining methods. 
In RIConv, the use of vector $\vv{pm}$ (from the representative point to the local centroid) as reference axis is unstable under noise. 
GCAConv and RI-GCN uses LRF which is also vulnerable to noise. 

\subsection{LRA/LRF Comparison}
\begin{figure}[t]
	\centering
	\includegraphics[width=0.90\linewidth]{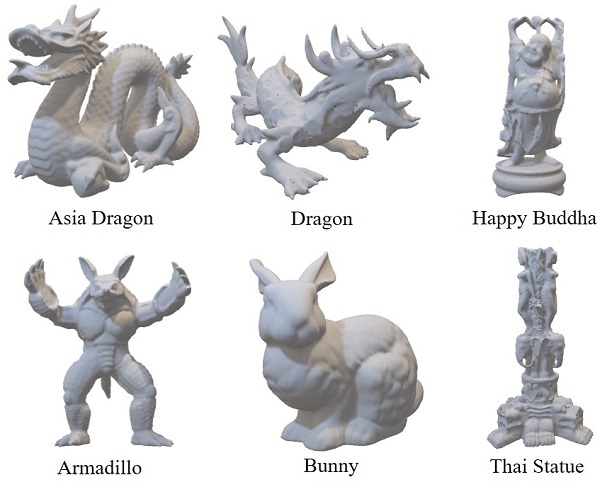}
	\\
	\vspace{0.2in}
	\includegraphics[width=0.9\linewidth]{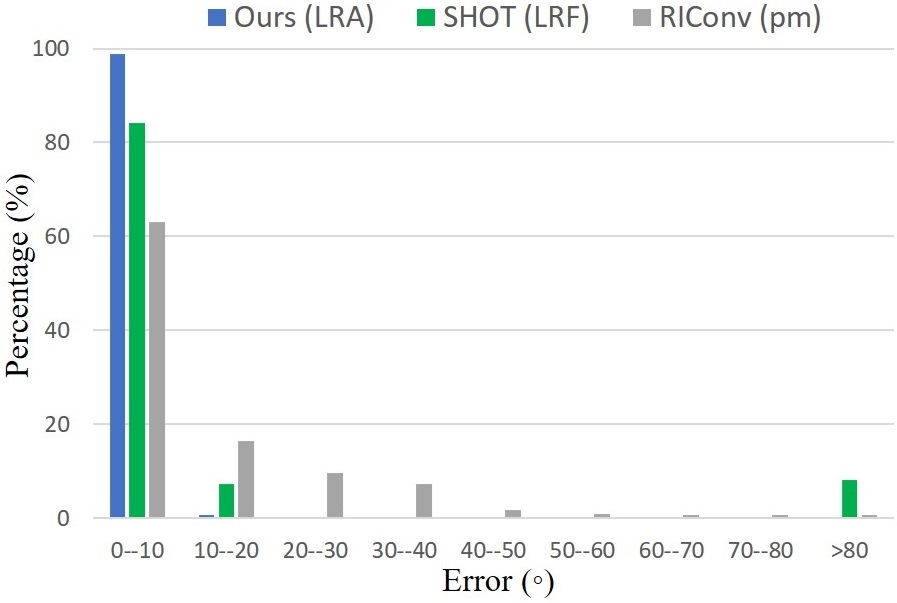}
	\caption{Repeatability of reference axis. We plot the histogram of errors of our LRA,  LRF~\citep{tombari2010unique}, and the $pm$ vector in RIConv~\citep{zhang-riconv-3dv19} for six models from the Stanford 3D Scanning Repository~\cite{curless1996volumetric}. LRA has the best lower range errors.}
	\label{fig:repeat}
\end{figure}

As the LRA is the basis to the define the Informative Rotation Invariant Features (IRIF), it is necessary to analyse the stability of LRA separately. A popular way to evaluate the robustness of LRA or LRF is to benchmark the repeatability.
We follow \citet{guo2013rotational} (see their section 3.3) to conduct this experiment. Noted that there are also methods that solve LRFs for mesh such as MeshHog~\citep{zaharescu2009surface} and RoPS~\citep{guo2013rotational}. In this study we assume no normal vectors or triangle faces so we omit such methods in our comparison.
We use six models from the Stanford 3D Scanning Repository~\citep{curless1996volumetric} (Figure~\ref{fig:repeat} top). The scenes are created by resampling the models down to 1/2 of their original mesh resolution with Gaussian noise added (0.5 mesh resolution).

From each model, 1000 points are randomly selected and their correspondences in the scene are obtained by searching the closest point in the Euclidean space. Let's denote the pair of points as $(p_{si}, p_{mi})$ from scene and model respectively. The LRFs or LRAs for these two points are computed as $V_{s_i}$ and $V_{m_i}$. To measure the similarity between them, we use the error evaluation metric provided by~\citet{mian2010repeatability}:
\begin{align}
e_{i} = \arccos\left(\frac{\mathrm{Tr}(V_{s_i}V_{m_i})-1}{2}\right)\frac{180}{\pi}.
\end{align}
Ideally, $e_{i}$ is zero when there is no error. 
We compare with LRF used in SHOT~\citep{tombari2010unique} and $pm$ vector used in RIConv~\citep{zhang-riconv-3dv19}.
The results are shown in Figure~\ref{fig:repeat}, where the horizontal axis indicates the angular error range and the vertical axis represents the percentage of points. The more points fall into left lower error range, the better of the methods. 
As can be seen, LRA have much more low-range angular errors than other methods, and has significantly less high-range errors.

\subsection{Ablation Studies}
\label{ablation}
\paragraph{IRIF Design.}
\newcommand{\tabincell}[2]{\begin{tabular}{@{}#1@{}}#2\vspace{-0.15in}\end{tabular}}
\begin{table}[t]
\caption{An evaluation of our features design. By combining distances and angles, relations between the representative points and the neighbors, relations among the neighbors, Model A with IRIF can achieve performance on ModelNet40 similar to PointCNN~\citep{li2018pointcnn} using xyz coordinates.}
\label{tab:ablation}
	\begin{center}
	\begin{tabularx}{\linewidth}{l|YYYY|YYYYY}
		\noalign{\smallskip}\hline\noalign{\smallskip}
		Model  &$\varphi$  & $\beta_0$ & $\beta_1$ & $\beta_2$ & $d$ &$\alpha_{0}$ &$\alpha_{1}$ &$\alpha_{2}$ & Acc. \\
		\noalign{\smallskip}\hline\noalign{\smallskip}		
		\tabincell{c}{A \\B \\ C \\ D} 
		& \tabincell{c}{\checkmark \\            \\ \checkmark \\            }
		& \tabincell{c}{\checkmark \\ \checkmark \\            \\            } 
		& \tabincell{c}{\checkmark \\ \checkmark \\            \\            } 
		& \tabincell{c}{\checkmark \\ \checkmark \\            \\            } 
		& \tabincell{c}{\checkmark \\            \\ \checkmark \\  \checkmark} 
		& \tabincell{c}{\checkmark \\ \checkmark \\   		   \\  \checkmark} 
		& \tabincell{c}{\checkmark \\ \checkmark \\            \\  \checkmark} 
		& \tabincell{c}{\checkmark \\ \checkmark \\            \\  \checkmark} 
		
		& \tabincell{c}{91.2\\ 90.5 \\ 83.3 \\ 88.4}\\ 
		
		\noalign{\smallskip}\hline
	\end{tabularx}
	\end{center}
\end{table}
We first experiment by turning on/off different rotation invariant components used in the IRIF construction (\ref{sec:irif}). 
The result of this experiment is shown in Table~\ref{tab:ablation}. 
Model A is our baseline setting with all rotation invariant features activated.
Model B has only angle features, and Model C has only distance features. 
From the comparison, we can see that turning off either feature types can deteriorate the results. 
When only distances are employed (Model C), the accuracy decreases to 83.3\%. 
In Model D, we only keep features on the radial direction with $d$, $\alpha_{0}$, $\alpha_{1}$ and $\alpha_{2}$ such that only the relations between the reference point and its neighbors are considered. This setting is used in RIConv~\citep{zhang-riconv-3dv19}. 
Compared to Model A, it shows that our proposed features is more effective than those by RIConv, and the improvement is explained by the additional consideration of the relations among the neighbor points.

\paragraph{Kernel Size.}
\begin{table}[t]
\caption{Performance comparisons with different kernel sizes. Kernel size $1$ has the best speed-accuracy trade-off.}
\label{tab:compare_kernel}
	\centering
	\footnotesize
	\begin{tabular}{lcccc}
		\noalign{\smallskip}\hline\noalign{\smallskip}
		Kernel Size   & Time / Epoch & Params. & Acc. \\
		\noalign{\smallskip}\hline\noalign{\smallskip}
        $1$   &40s   & 0.40M & 91.2 \\
		$3$   &63s  & 0.83M & 91.2 \\
		$5$   &73s & 1.29M & 91.0 \\
		$7$   &85s & 1.73M & 91.0 \\
		\noalign{\smallskip}\hline
	\end{tabular}
\end{table}
For the pointwise convolution in Algorithm~\ref{alg:conv}, different kernel sizes can be chosen. 
We compare the performance with different kernel sizes as shown in Table~\ref{tab:compare_kernel}. 
With larger kernels, the running time and the number of parameters also increases. 
We observed that larger kernels ($5$ and $7$) do not correspond to better performance because our rotation-invariant features only involves a point $x_i$ and its immediate neighbors, which makes a kernel up to size $3$ sufficient to capture the local features.
In our experiments, we choose kernel size $1$ for the best speed and accuracy trade-off. 

\paragraph{IRIF on different network architectures}
IRIF can be applied for other architectures to realize rotation invariance. Here, we experiment with PointNet++~\citep{qi2017pointnet++} and DGCNN~\citep{wang2018edgeconv} by replacing their xyz coordinates input with IRIF. The results are shown in Table~\ref{tab:ablation_irif_network}. We can see that rotation invariance can be achieved.

\begin{table}[t]
\caption{Application of IRIF features on different network architecture.}
\label{tab:ablation_irif_network}
	\centering
	\footnotesize
	\begin{tabular}{l|ccc|c}
		\noalign{\smallskip}\hline\noalign{\smallskip}
		Method   & z/z & SO3/SO3 & z/SO3 &Acc. Std\\
		\noalign{\smallskip}\hline\noalign{\smallskip}
		PN++    &89.3 &85.0 &28.6 &33.8\\
		PN++ (IRIF)   &89.1 &89.1 &89.0 &\textbf{0.0} \\
		\noalign{\smallskip}\hline\noalign{\smallskip}
		DGCNN   &92.2 &81.1 &20.6  &38.5\\
		DGCNN  (IRIF)     &90.2 &90.1 &90.2 &\textbf{0.0} \\
		\noalign{\smallskip}\hline
	\end{tabular}
\end{table}

\begin{figure}[t]
	\centering
	\includegraphics[width=0.90\linewidth]{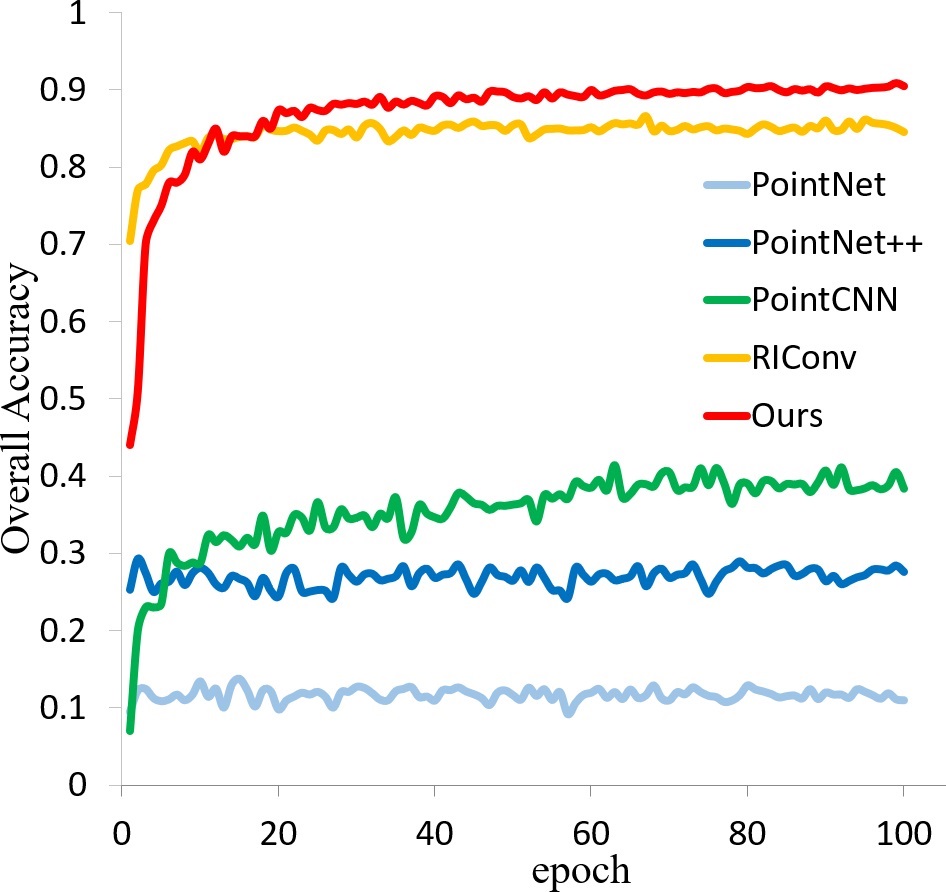}
	\caption{Overall accuracy versus training epochs plot of object classification under z/SO3 mode on the ModelNet40 dataset.}
	\label{fig:training}
\end{figure}

\paragraph{Training Statistics.}
We further analyze the training convergence and measure the accuracy on the test set on different training epochs. 
We use the z/SO3 setting. 
The plot is shown in Figure~\ref{fig:training}. It can be seen that RIConv~\citep{zhang-riconv-3dv19} converges to a flat accuracy curve after about 20 epochs. Our new method converges with higher accuracy, and the accuracy still increases by 100 epochs. Both methods outperform standard point cloud methods~\citep{qi2017pointnet,qi2017pointnet++,li2018pointcnn} by a large margin as expected for the z/SO3 setting. 

\subsection{Real World 3D Point Cloud Classification}
\label{sec:scanobj}
\begin{table*}[t]
	\centering
	\caption{Comparisons of real world 3D point cloud classification on ScanObjectNN~\citep{uy-scanobjectnn-iccv19} dataset. We tested on the OBJ\_ONLY (object without background), OBJ\_BG (object with background) and PB\_T50\_RS (hardest) variant.}
    \label{tab:classification_scanobjectnn}
	\begin{tabular}{l|ccc|ccc|ccc}
		\noalign{\smallskip}\hline\noalign{\smallskip}
		&             &OBJ\_ONLY      &    &    &OBJ\_BG   &     &        &PB\_T50\_RS  &\\
		Method  &z/z  &SO3/SO3 &z/SO3  &z/z  &SO3/SO3 &z/SO3      &z/z  &SO3/SO3 &z/SO3\\
		\noalign{\smallskip}\hline\noalign{\smallskip}
		PointNet~\citep{qi2017pointnet}      &79.2 &57.5 &28.7 &73.3 &54.7 &16.7 &68.2 &42.2 &17.1\\
		PointNet++~\citep{qi2017pointnet++}  &84.3 &57.1 &25.6 &82.3 &47.4 &15.0 &77.9 &60.1 &15.8\\
		PointCNN~\citep{li2018pointcnn}      &85.5 &66.9 &21.9  &\textbf{86.1} &63.7 &14.6 &78.5 &51.8 &14.9\\
		DGCNN~\citep{wang2018edgeconv}       &86.2 &73.6 &20.7 &82.8 &71.8 &17.7 &78.1 &63.4 &16.1\\
		\noalign{\smallskip}\hline\noalign{\smallskip}
		RIConv ~\citep{zhang-riconv-3dv19}   &79.8 &79.8 &79.8 &78.4 &78.2 &78.4 &68.1 &68.3 &68.3\\
		GCAConv~\citep{zhang2020global}      &80.1 &80.3 &80.1 &78.2 &78.1 &78.2 &69.8 &70.0 &69.8\\
		\noalign{\smallskip}\hline\noalign{\smallskip}
		Ours (w/o normal) &\textbf{86.2} &\textbf{86.2} &\textbf{86.2} &85.6 &\textbf{85.6} &\textbf{85.6} &\textbf{80.3} &\textbf{80.3} &\textbf{80.3}\\
		\noalign{\smallskip}\hline
	\end{tabular}
\end{table*}

\begin{table*}[t]
	\centering
	\caption{Comparisons of 3D shape retrieval on the ShapeNet Core~\citep{wu20153d}. The accuracy (\%) is reported based on the standard evaluation metrics including precision, recall, f-score, mean average precision (mAP) and normalized discounted cumulative gain (NDCG).}
    \label{tab:retrieval}
    \resizebox{\textwidth}{!}{%
	\begin{tabular}{l|ccccc|ccccc|c}
		\noalign{\smallskip}\hline\noalign{\smallskip}
		&   &     &micro &   &      &   &    &macro&    &     & \\
		Method &PN &R@N  &F1@N &mAP & NDCG &PN &R@N &F1@N &mAP &NDCG & Score\\
		\noalign{\smallskip}\hline\noalign{\smallskip}
		\citet{furuya2016deep}  &81.4 &68.3 &70.6 &65.6 &75.4 &60.7 &53.9 &50.3 &47.6 &56.0 &56.6\\
		\citet{tatsuma2009multi}  &70.5 &76.9 &71.9 &69.6 &78.3 &42.4 &56.3 &43.4 &41.8 &47.9 &55.7 \\
		\citet{bai2016gift}   &66.0 &65.0 &64.3 &56.7 &70.1 &44.3 &50.8 &43.7 &40.6 &51.3 &48.7\\
		\noalign{\smallskip}\hline\noalign{\smallskip}
		\citet{esteves2018learning}  &71.7 &73.7 &- &68.5 &- &45.0 &55.0 &- &44.4 &- &56.5\\
		SFCNN~\citep{rao-spherical-cvpr19}  &77.8 &75.1 &75.2 &70.5 &81.3 &65.6 &53.9 &53.6 &48.3 &58.0 &59.4\\
		GCAConv~\citep{zhang2020global} &82.9 &76.3 &74.8 &70.8 &81.3 &66.8 &55.9 &51.2 &49.0	&58.2	&\textbf{61.2}\\
		RIF~\citep{li2021rotation} &82.1 &73.7 &74.1 &70.7 &80.5 &51.2 &\textbf{66.4} &\textbf{55.8} &\textbf{51.0} &56.0 &60.9\\
		\noalign{\smallskip}\hline\noalign{\smallskip}
		Ours &\textbf{83.2} &\textbf{77.2} &\textbf{75.2} &\textbf{71.3} &\textbf{81.8} &\textbf{68.1} &58.2 &53.7 &50.2	&\textbf{58.8}	&60.7\\
		\noalign{\smallskip}\hline
	\end{tabular}
	}
\end{table*}

Real-world point cloud data usually contains missing data, occlusions, and non-uniform density. We evaluate the classification performance on ScanObjectNN~\citep{uy-scanobjectnn-iccv19}, a real-world dataset of 3D point clouds captured by RGB-D camera. 
The data comprises 2902 objects classified into 15 categories sampled from real-world indoor scenes.
For our evaluation, we use the processed files and choose the easiest variant OBJ\_ONLY and OBJ\_BG (only object without/with background points, without rotation, translation, and scaling) and the hardest variant PB\_T50\_RS (with 50\% bounding box translation, rotation around the gravity axis, and random scaling that result in rotated and partial data).
The results are shown in Table~\ref{tab:classification_scanobjectnn}. 
It can be seen that our method outperforms other existing approaches across all the scenarios and only slightly worse than PointCNN under the z/z scenario. 
This verifies that \ourconv is also effective on real-world datasets. Note that we only test the 'w/o normal' case as the normal vectors are not provided in the processed files of this dataset.

The experiment with ScanObjectNN also demonstrates the robustness of our method in the presence of occlusion or background points. 
Firstly, we can see that our method works well for objects in ScanObjectNN despite that they are incomplete scans due to occlusions, which demonstrates the effectiveness of our rotation-invariant features (IRIF) and the construction of local reference axes (LRA). 
Secondly, when additional background points are added to the objects as demonstrated in the OBJ\_BG variant, comparing between OBJ\_ONLY and OBJ\_BG, we observe that our method is more tolerant and has less accuracy drop ($-0.6\%$) compared to previous methods (RIConv with $-1.4\%$ drop and GCAConv with $-1.9\%$ drop in z/z case).

\subsection{Shape Retrieval}
\label{retrieval}
Another popular task to evaluate rotation invariance on 3D shape is shape retrieval~\citep{savva2016shrec16}. 
Here we conducted experiments on ShapeNet Core~\citep{wu20153d}, following the perturbed protocol of the SHREC'17 3D shape retrieval contest~\citep{savva2016shrec16} and the experiment setting of SFCNN~\citep{rao-spherical-cvpr19}. 
We use the same output features from the bottleneck layer in the network (similar to features used in the classification task; see Figure~\ref{fig:rinet}).  
We compare our method with methods proposed in SHREC’17~\citep{furuya2016deep,tatsuma2009multi,bai2016gift} and two recent methods on rotation-invariant convolution~\citep{esteves2018learning,rao-spherical-cvpr19}. 
The results are shown in Table~\ref{tab:retrieval}.
Similar to the classification task, our method achieves the state-of-the-art result, outperforming previous methods for most evaluation metrics.

\subsection{Object Part Segmentation on ShapeNet}
\label{seg_shapenet}
We also evaluated our method with the object part segmentation where each point of the input point cloud is predicted with a part label.
We train and test with the ShapeNet dataset~\citep{chang2015shapenet} that contains $16,880$ CAD models in $16$ categories. 
Each model is annotated with $2$ to $6$ parts, resulting in a total of $50$ object parts. 
We follow the standard train/test split with $14,006$ models for training and $2,874$ models for testing, respectively. 

\begin{figure}[t]
\centering
\includegraphics[width=\linewidth]{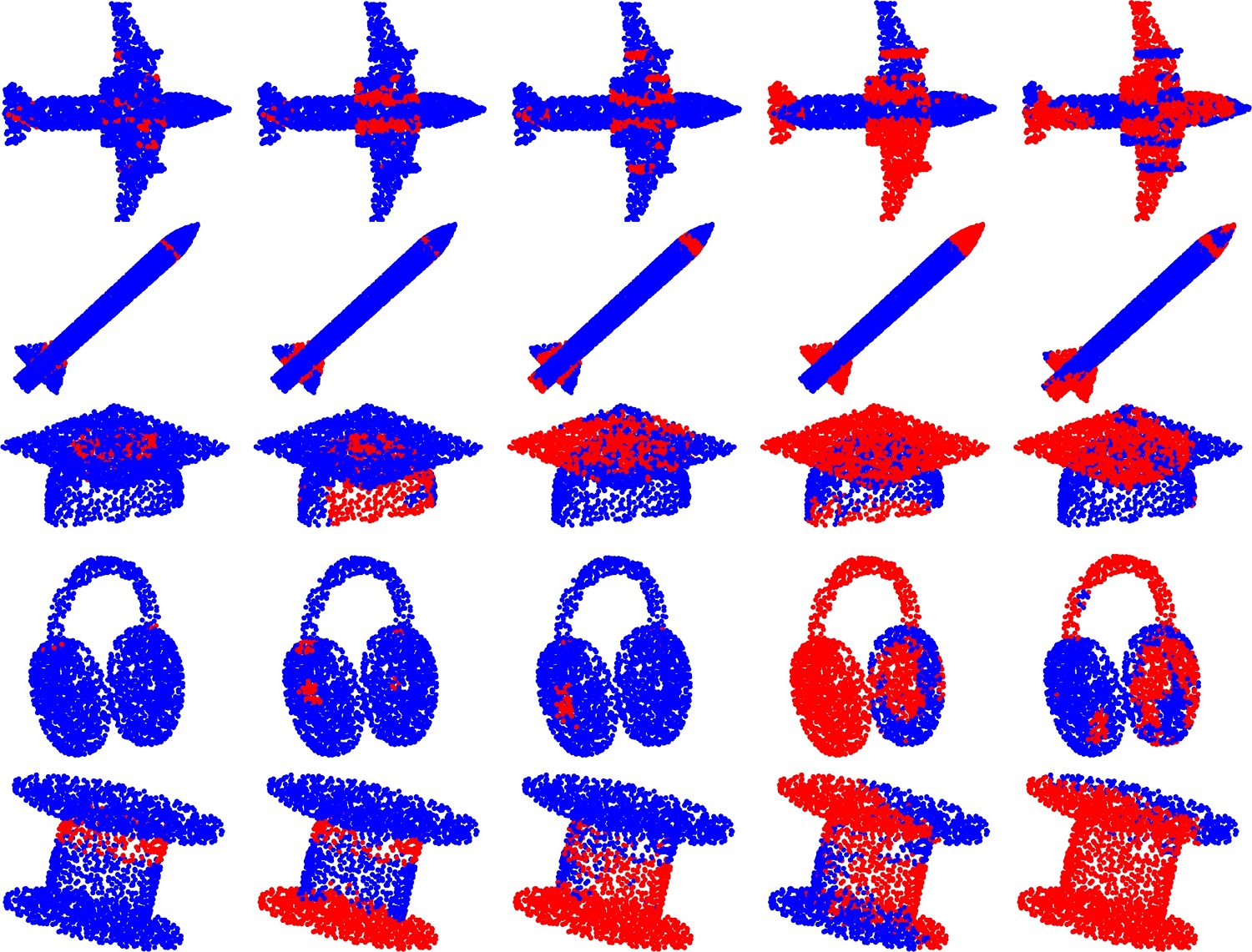}
\parbox[b]{1\linewidth}{\relax
		~~  Ours ~~~~~ GCAConv ~~~~ RIConv \quad PointNet++ \quad SpiderCNN}
\caption{Comparison of part segmentation error maps on ShapeNet dataset under z/SO3 mode. The red points indicate wrong segmentations.}
\label{fig:err_map}
\end{figure}

The evaluation results are shown in Table~\ref{tab:partsegmentation}.  Our method outperforms translation-invariant convolution methods significantly in the SO3/SO3 and z/SO3 scenario. 
Compared to previous rotation-invariant convolution methods, our method also has better performance. 
This result aligns well with the performance reported in the object classification task. 
Our method also has consistent performance for both rotation cases, which empirically confirms the rotation invariance in our convolution.
We illustrate the qualitative results by error maps as shown in Figure~\ref{fig:err_map}. 
By comparing with the groundtruth, we plot the wrong segmentation points as red. 
The plot clearly shows that our predictions are the closest to the ground truth. 

\begin{table}[t]
\centering
\caption{Comparisons of object part segmentation performed on ShapeNet dataset~\citep{chang2015shapenet}. The mean per-class IoU (mIoU, \%) is used to measure the accuracy under two challenging rotation modes: SO3/SO3 and z/SO3.}
\label{tab:partsegmentation}
\begin{tabular}{l|Hcc} 
\hline\noalign{\smallskip}
Method   & Input & SO3/SO3 & z/SO3  \\
\noalign{\smallskip}\hline\noalign{\smallskip}
PointNet~\citep{qi2017pointnet}     & xyz        & 74.4   & 37.8  \\
PointCNN~\citep{li2018pointcnn}     & xyz        & 71.4   & 34.7  \\
DGCNN~\citep{wang2018edgeconv}      & xyz        & 73.3   & 37.4  \\
RS-CNN~\citep{liu2019relation}      & xyz        & 72.5   & 36.5  \\
\noalign{\smallskip}\hline\noalign{\smallskip}
RIConv~\citep{zhang-riconv-3dv19}  & xyz        & 75.5   & 75.3  \\
GCAConv~\citep{zhang2020global}     & xyz        & 77.3   & 77.2  \\
RI-GCN~\citep{kim2020rotation}      & xyz       & 77.0   & 77.0 \\
RIF~\citep{li2021rotation}   & xyz  & 79.4  & 79.2\\
Ours (w/o normal)             & xyz &  \textbf{80.3} &  \textbf{80.3} \\
\noalign{\smallskip}\hline\noalign{\smallskip}
PointNet++~\citep{qi2017pointnet++} & xyz+nor & 76.7   & 48.2  \\
SpiderCNN~\citep{xu2018spidercnn}   & xyz+nor & 72.3   & 42.9  \\
Ours (w/ normal)           & xyz+nor &  \textbf{80.5} &  \textbf{80.5} \\
\noalign{\smallskip}\hline
\end{tabular}
\end{table}

\subsection{Large-scale scene segmentation on S3DIS}
We conduct an experiment for the scene segmentation on the S3DIS dataset~\citep{armeni20163d} which is a large-scale indoor scene dataset comprising 3D scans from Matterport scanners in 6 indoor areas including 271 rooms with each point is annotated with one of the semantic labels from 13 categories. In this experiment, we use Area-5 for testing to better measure the
generalization ability, and report the results in SO3/SO3 and z/SO3 scenario.
The results are shown in Table~\ref{tab:s3dis}. 
From the results, we can see that RIConv++ outperforms RIConv~\citep{zhang-riconv-3dv19} and GCAConv~\citep{zhang2020global}, and works consistently for both scenarios.

\begin{table}[t]
\caption{Comparisons of the semantic segmentation accuracy (mIoU, \%) on the S3DIS dataset (Area-5).}
\label{tab:s3dis}
	\centering
	\begin{tabular}{l|cc}
		\noalign{\smallskip}\hline\noalign{\smallskip}
		Method   & SO3/SO3 & z/SO3\\
		\noalign{\smallskip}\hline\noalign{\smallskip}
		PointNet~\citep{qi2017pointnet}   & 35.2 & 20.8 \\
		PointCNN~\citep{li2018pointcnn}   & 43.5 & 23.6 \\
		\noalign{\smallskip}\hline\noalign{\smallskip}
		RIConv ~\citep{zhang-riconv-3dv19}  &53.3 &53.2 \\
		GCAConv~\citep{zhang2020global}     &55.8 &55.7 \\
		Ours (w/o normal)    & \textbf{57.0} & \textbf{57.1}\\
		\noalign{\smallskip}\hline
	\end{tabular}
\end{table}

\section{Conclusion}
\label{conclusion}
In this work, we revisited the design of rotation-invariant features for 3D point cloud convolution, and proposed \ourconv for 3D point cloud convolution that satisfies rotation invariance property. 
The effectiveness of \ourconv comes from the powerful and stable rotation invariant features and the flexible neighborhood size employed in the convolution. 
We used local reference axis (LRA) instead of the commonly used local reference frame (LRF) to construct stable rotation invariant features. 
The relationship between the representative point and its neighbors and relationship among neighbors are used in tandem to make our convolution much informative. 
The local-global information can be captured by simply using different neighborhood size in each convolution.  
Such a design achieves the state-of-the-art results in various tasks such as object classification, shape retrieval, and object part segmentation. 

As our newly proposed convolution can closely match the performance of state-of-the-art translation-invariant convolutions, our work opens up opportunities to further reduce the performance gap between rotation- and translation-invariant convolution especially in the presence of noise. 
Our method also shows that handcrafted rotation-invariant features, when used properly, can lead to compelling results in 3D deep learning. There remains an open question: it would be of great interest to design a convolution that can learn rotation-invariant features automatically, without the need of handcrafted features, which could allow the applications of rotation-invariant features to more data domains.

\begin{acknowledgements}
We thank the anonymous reviewers for their constructive comments. 
This research project is supported by the grant from Ningbo Research Institute of Zhejiang University (1149957B20210125), and partially supported by an internal grant from HKUST (R9429).
\end{acknowledgements}

%
%


\bibliographystyle{spbasic}      
\bibliography{egbib}   

\end{document}